%% file: colm2024_conference.tex
\documentclass{article} 
\usepackage{colm2024_conference}

\usepackage{microtype}
\usepackage{hyperref}
\usepackage{url}
\usepackage{booktabs}
\usepackage{enumitem}
\usepackage{graphicx}
\usepackage{multirow}
\usepackage{pifont}
\usepackage{xcolor}
\usepackage{booktabs}
\usepackage{hyperref}
\usepackage{makecell}
\usepackage{amsmath}
\usepackage{amsfonts}
\usepackage{enumitem}
\usepackage{multicol}
\usepackage{soul}
\usepackage{arydshln}
\usepackage{subcaption}
\usepackage{wrapfig}

\usepackage{color}  

\newcommand{\bench}{\texttt{VisualWebBench}}

\usepackage{color-edits}
\addauthor{gn}{magenta}
\addauthor{xiang}{blue}
\addauthor{yuchen}{purple}

\title{\bench: How Far Have Multimodal LLMs Evolved in Web Page Understanding and Grounding?}

\author{Junpeng Liu$^{\clubsuit,*}$~~Yifan Song$^{\circ,*}$~~Bill Yuchen Lin$^{\mathsection}$~~Wai Lam$^{\clubsuit}$~~Graham Neubig$^{\spadesuit}$\\
\textbf{Yuanzhi Li$^{\diamondsuit}$~~Xiang Yue$^{\spadesuit}$}\\[10pt]
$^{\spadesuit}$ Carnegie Mellon University \\
$^{\clubsuit}$ The Chinese University of Hong Kong \\
$^{\circ}$ School of Computer Science, Peking University \\
$^{\diamondsuit}$ MBZUAI~~ 
$^{\mathsection}$ Allen Institute for AI 
}

\colmfinalcopy 
\begin{document}

\maketitle

\vspace{-0.7cm}
\begin{center}
    \url{https://visualwebbench.github.io/}
\end{center}
\vspace{0.2cm}
\vspace{5pt}

\renewcommand{\thefootnote}{\fnsymbol{footnote}}
    \footnotetext[1]{Equal contribution.}
    \footnotetext[2]{Corresponding to: {\texttt{jpliu@link.cuhk.edu.hk~~xyue2@andrew.cmu.edu}}.}
\renewcommand{\thefootnote}{\arabic{footnote}}

\begin{abstract}
Multimodal Large Language models (MLLMs) have shown promise in web-related tasks, but evaluating their performance in the web domain remains a challenge due to the lack of comprehensive benchmarks. Existing benchmarks are either designed for general multimodal tasks, failing to capture the unique characteristics of web pages, or focus on end-to-end web agent tasks, unable to measure fine-grained abilities such as OCR, understanding, and grounding. In this paper, we introduce \bench{}, a multimodal benchmark designed to assess the capabilities of MLLMs across a variety of web tasks. \bench{} consists of seven tasks, and comprises 1.5K human-curated instances from 139 real websites,  covering 87 sub-domains. We evaluate 14 open-source MLLMs, Gemini Pro, Claude-3 series, and GPT-4V(ision) on \bench{}, revealing significant challenges and performance gaps. Further analysis highlights the limitations of current MLLMs, including inadequate grounding in text-rich environments and subpar performance with low-resolution image inputs. We believe \bench{} will serve as a valuable resource for the research community and contribute to the creation of more powerful and versatile MLLMs for web-related applications.

\end{abstract}

\input{1-intro-v2}
\input{4-related}

\input{2-bench}
\input{3-experiment}

\input{9-conclusion}

\section*{Acknowledgement}
The authors would thank Boyuan Zheng, Shuyan Zhou, Yizhong Wang, and Jie Huang for their insightful discussions and comments. The authors would also thank seven annotators for their help in annotating the action grounding task samples. 

\bibliography{colm2024_conference}
\bibliographystyle{colm2024_conference}

\clearpage
\appendix

\section{Annotation Tool of Action Grounding}
\label{sec:annotation_tool}
We developed an annotation tool to facilitate the annotation of the action grounding task. 
The annotation procedure is as follows:
\begin{enumerate}[leftmargin=*, nolistsep]
\setlength{\itemsep}{1mm}
\item Learn about what the shown website is for, based on the presented website descriptions, and you may still need to search for the website name in Google to have a better understanding.
\item Refer to action description examples generated by GPT-4V, and then write your instruction. Then, click ``Confirm instruction''. Please make your instructions diverse, and do not write too many instructions like ``search for an item''.
\item Move the Mouse to hover over the corresponding element that will be interacted with to accomplish the action description, then press key ``s'' (instead of CLICK) to select it. After that, a green rectangle will be shown to indicate the selected element. Note that the element should be interactive (e.g.,  clickable or inputtable, etc. ). 
\item Confirm that the selected element (indicated by a blinking green rectangle) correctly corresponds to the action description and click the "submit" button, then click "allow" to allow screen capture.
\end{enumerate}

\begin{figure}[ht]
    \centering
    \includegraphics[width=0.8\textwidth]{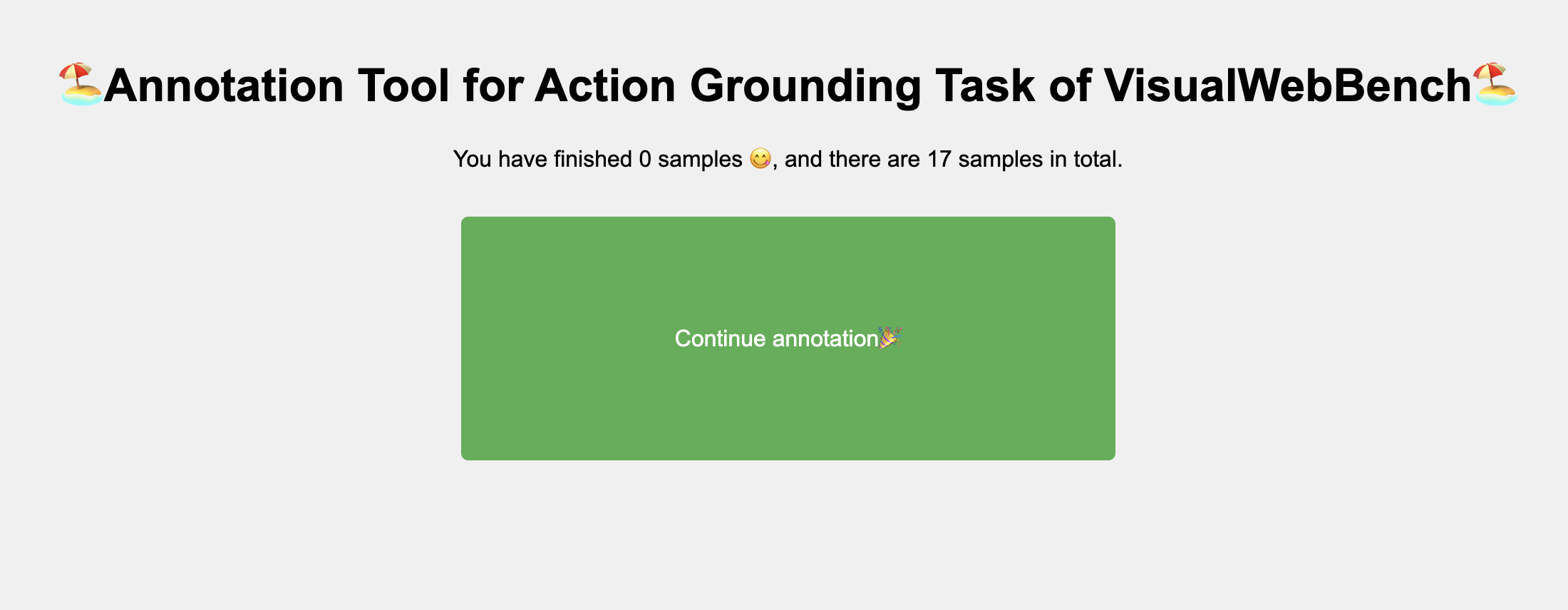}
    \caption{Illustration of the annotation tool (1).}
    \label{fig:anno_tool1}
\end{figure}

\begin{figure}[ht]
    \centering
    \includegraphics[width=0.8\textwidth]{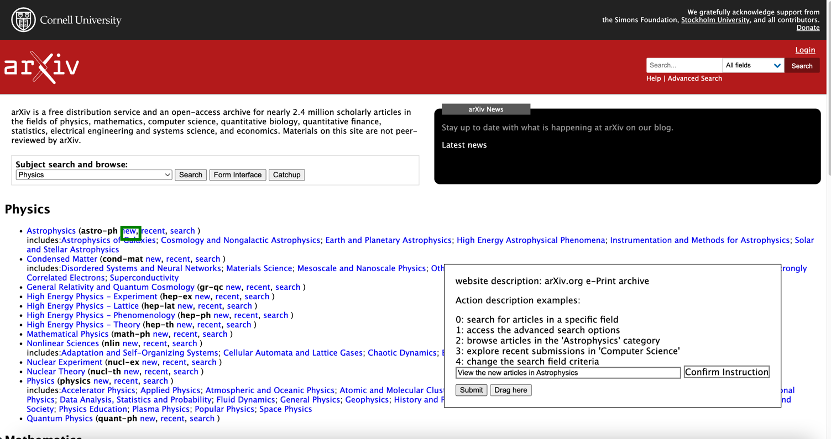}
    \caption{Illustration of the annotation tool (2).}
    \label{fig:anno_tool2}
\end{figure}

\section{Data Verification and Curation}
\label{sec:curation}
All samples of our benchmark undergo careful verification and curation through a collaborative effort and a division of tasks by two authors. The process encompasses: 
\begin{itemize}[leftmargin=*, nolistsep]
\setlength{\itemsep}{1mm}
\item Ensuring the main content within screenshots remains unobscured by advertisements or intrusive banners.
\item Verifying that the captions describe the most important information generally of the websites for Captioning.
\item Headings of websites are correctly extracted for Heading OCR.
\item The annotated bounding boxes properly encapsulate the target web element description for Element OCR and Element Grounding.
\item The annotated bounding boxes are well aligned with the title of redirected websites for Action Prediction.
\item The instructions are appropriately matched with their annotated bounding boxes for Action Grounding.
\end{itemize}

\section{Details of Evaluated MLLMs}
\label{sec:experiment-lmms}

We consider various general large multimodal models.
By default, for each model family, we use the latest, largest, and best-performing available checkpoint to date.
\textit{(i)} BLIP-2~\citep{li2023blip} series bridges the vision-language modality gap with a lightweight Q-Former.
\textit{(ii)} InstructBLIP~\citep{dai2024instructblip} further performs vision-language instruction tuning based BLIP-2 models.
\textit{(iii)} mPLUG-Owl2~\citep{ye2023mplug} adapts a modularized network to facilitate modality collaboration while preserving specific features.
\textit{(iv)} Otter~\citep{li2023mimic} has improved instruction following and in-context learning capabilities.
\textit{(v)} VILA~\citep{lin2023vila} is pretrained with interleaved image-text data at scale.
\textit{(vi)} Fuyu~\citep{fuyu-8b} is a decoder-only transformer and treats image tokens like text tokens.
\textit{(vii)} SPHINX~\citep{lin2023sphinx} mixes different tuning tasks, and visual embeddings to build a versatile MLLM.
\textit{(viii)} LLaVA-1.5~\citep{liu2024visual} combines a vision encoder and Vicuna for general-purpose visual and language understanding, and LLaVA-1.6~\citep{liu2024llavanext} family is the enhanced version with improved image resolution, reasoning, OCR, and world knowledge.
We consider three scales: Vicuna-7B, Vicuna-13B, and Hermes-Yi-34B for model scaling analysis.
\textit{(ix)} Qwen-VL~\citep{bai2023qwen} introduces trainable query embeddings and single-layer cross-attention module to bridge the modalities.
\textit{(x)} DeepSeek-VL~\citep{lu2024deepseek} incorporates a hybrid vision encoder to processe high-resolution images.
\textit{(xi)} Yi-VL~\citep{young2024yi} connects the vision encoder with MLLM with a simple MLP projection module and undergoes a three-stage training process.
\textit{(xii)} CogVLM~\citep{wang2023cogvlm} bridges the modality gap by a trainable visual expert module in the attention and FFN layers of the transformer.
We also include Gemini Pro~\citep{team2023gemini}, Claude Sonnet, Claude Opus~\citep{claude}, and GPT-4V(ision)~\citep{2023GPT4VisionSC} for comparison.

For all MLLMs, we set the temperature to 0.0 for deterministic generations.
All experiments are conducted on NVIDIA A100 80G GPUs.


\section{\bench{} vs. Mind2Web}
\label{sec:webagent}
Table~\ref{tab:webben_and_agent} details the scores of MLLMs on \bench{} and Mind2Web. 

\begin{table}
\centering
\begin{tabular}{lcccc}
\toprule
& VisualWebBench & Mind2Web \\
\midrule
SeeClick      & 9.7  & 20.9* \\
Qwen-VL       & 23.9 & 10.2* \\
CogAgent      & 28.7 & 15.5\dag \\
LLaVA-1.5-7B  & 17.0 & 4.0  \\
LLaVA-1.5-13B & 19.4 & 9.6  \\
LLaVA-1.6-7B  & 36.0 & 3.6  \\
LLaVA-1.6-13B & 39.4 & 6.3  \\
LLaVA-1.6-34B & 50.5 & 13.6 \\
Gemini-Pro    & 48.0 & 17.7\dag \\
GPT-4V(ison)  & 64.6 & 36.5\dag \\
\bottomrule
\end{tabular}
\caption{The comparison between scores of \bench{} and Mind2Web. * indicates the results taken from \citet{cheng2024seeclick}, while \dag \ denotes those taken from \citet{zheng2024gpt}. }
\label{tab:webben_and_agent}
\end{table}



\section{Additional Case Studies}
\label{sec:case_webqa}
\begin{figure}
    \centering
    \includegraphics[width=0.65\textwidth]{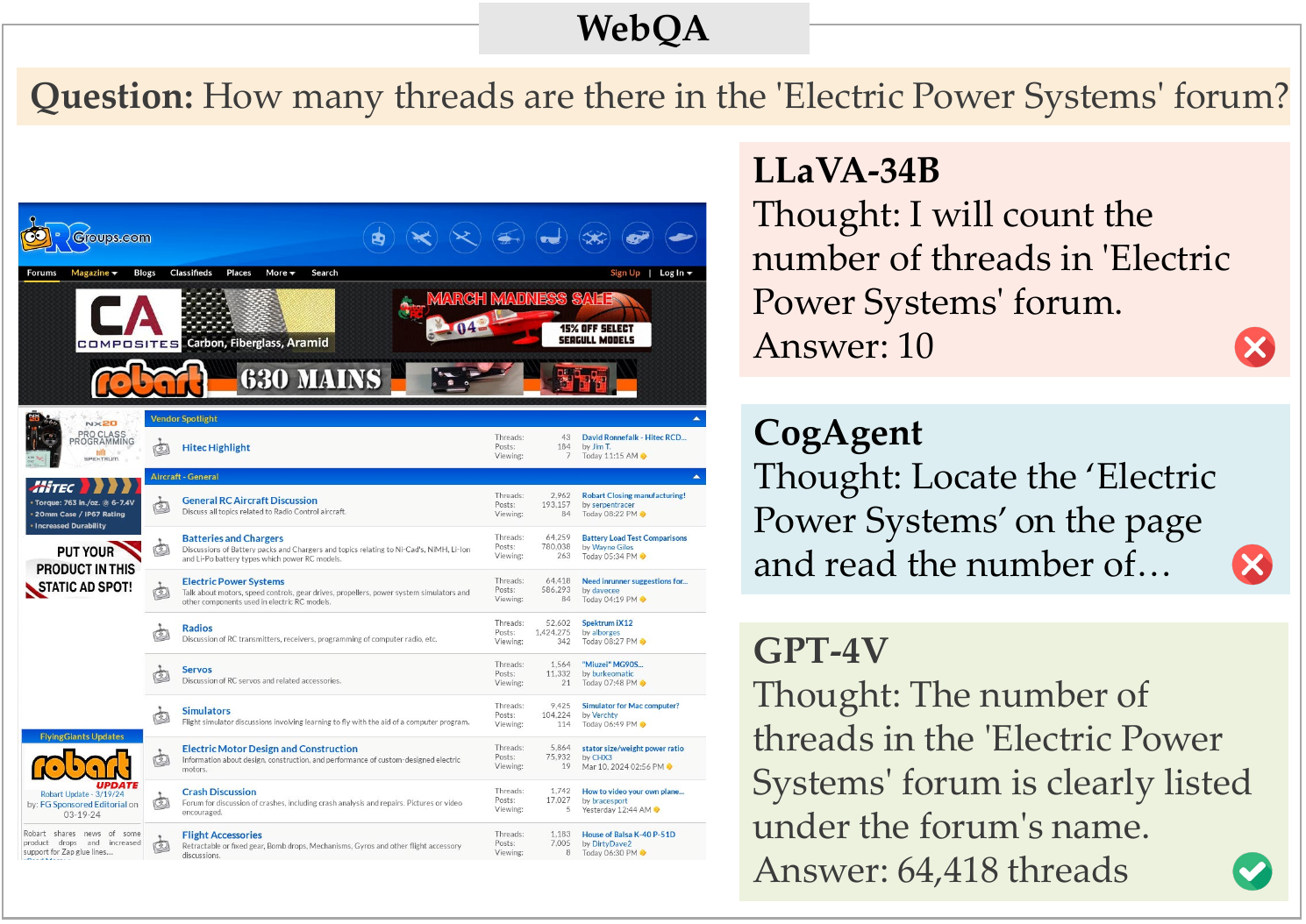}
    \caption{Case study of WebQA. }
    \label{fig:case_webqa}
\end{figure}

In Figure~\ref{fig:case_webqa}, 
CogAgent cannot understand the instruction correctly and output unrelated content. LLaVA-34b captures the instruction, yet generates a wrong answer. In contrast, both the thinking process and the answer of GPT-4V are correct. This demonstrates the strong understanding and reasoning ability of GPT-4V.

\begin{figure}
    \centering
    \includegraphics[width=0.65\textwidth]{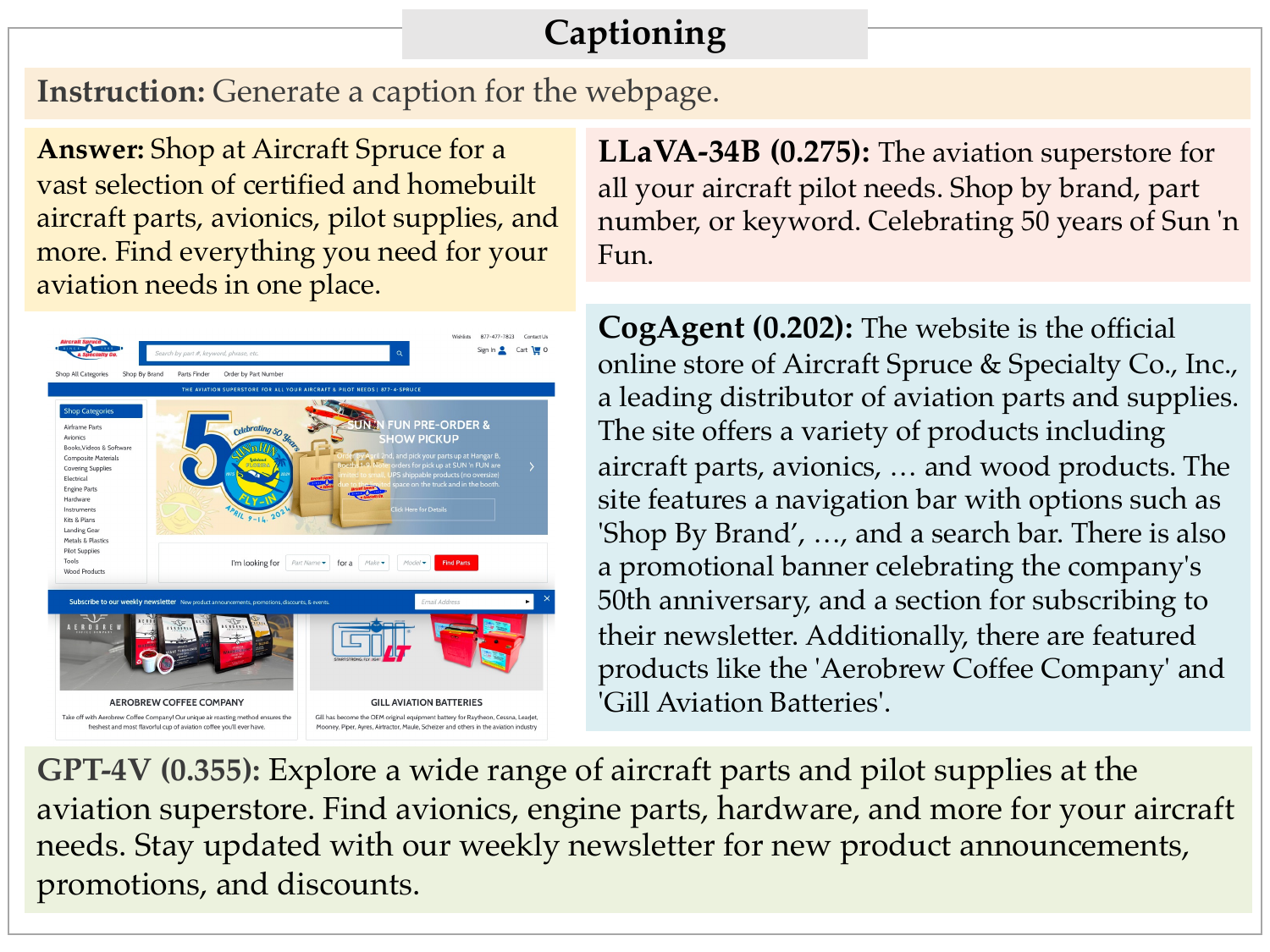}
    \caption{Case study of Captioning. }
    \label{fig:case_caption}
\end{figure}

\begin{figure}
    \centering
    \includegraphics[width=0.65\textwidth]{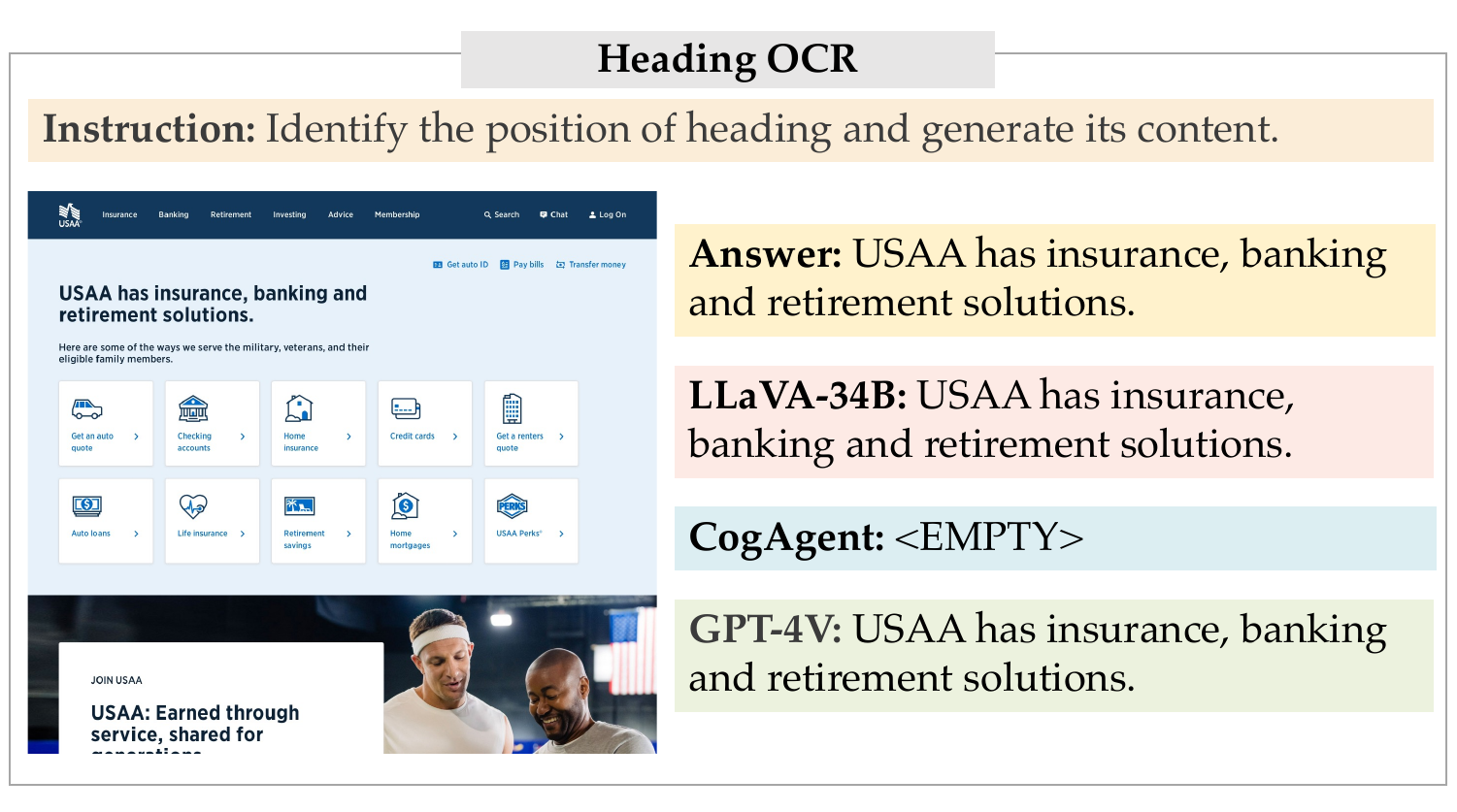}
    \caption{Case study of Heading OCR. }
    \label{fig:case_heading_ocr}
\end{figure}

\begin{figure}
    \centering
    \includegraphics[width=0.65\textwidth]{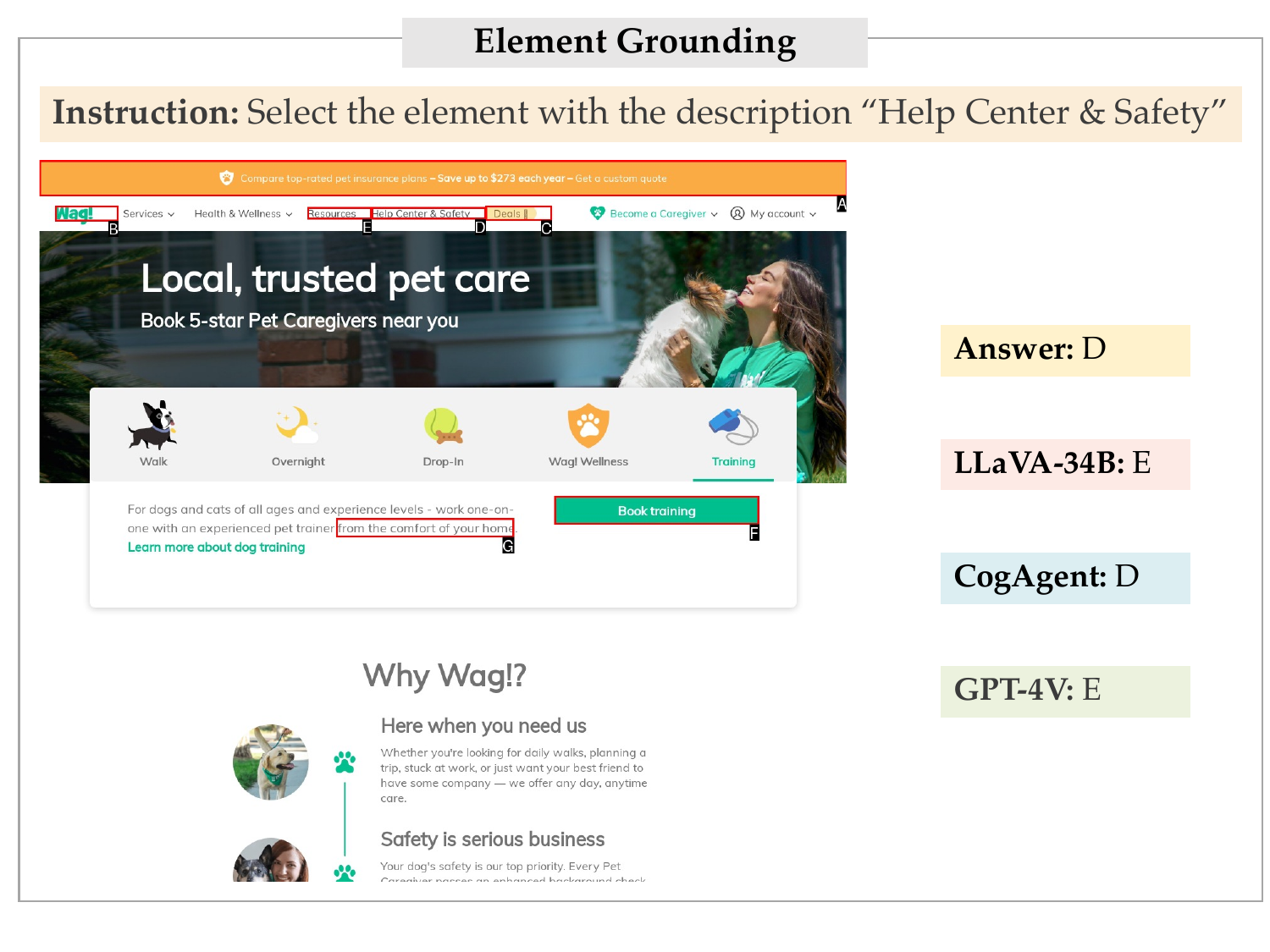}
    \caption{Case study of Element Grounding. }
    \label{fig:case_element_grounding}
\end{figure}

\begin{figure}
    \centering
    \includegraphics[width=0.65\textwidth]{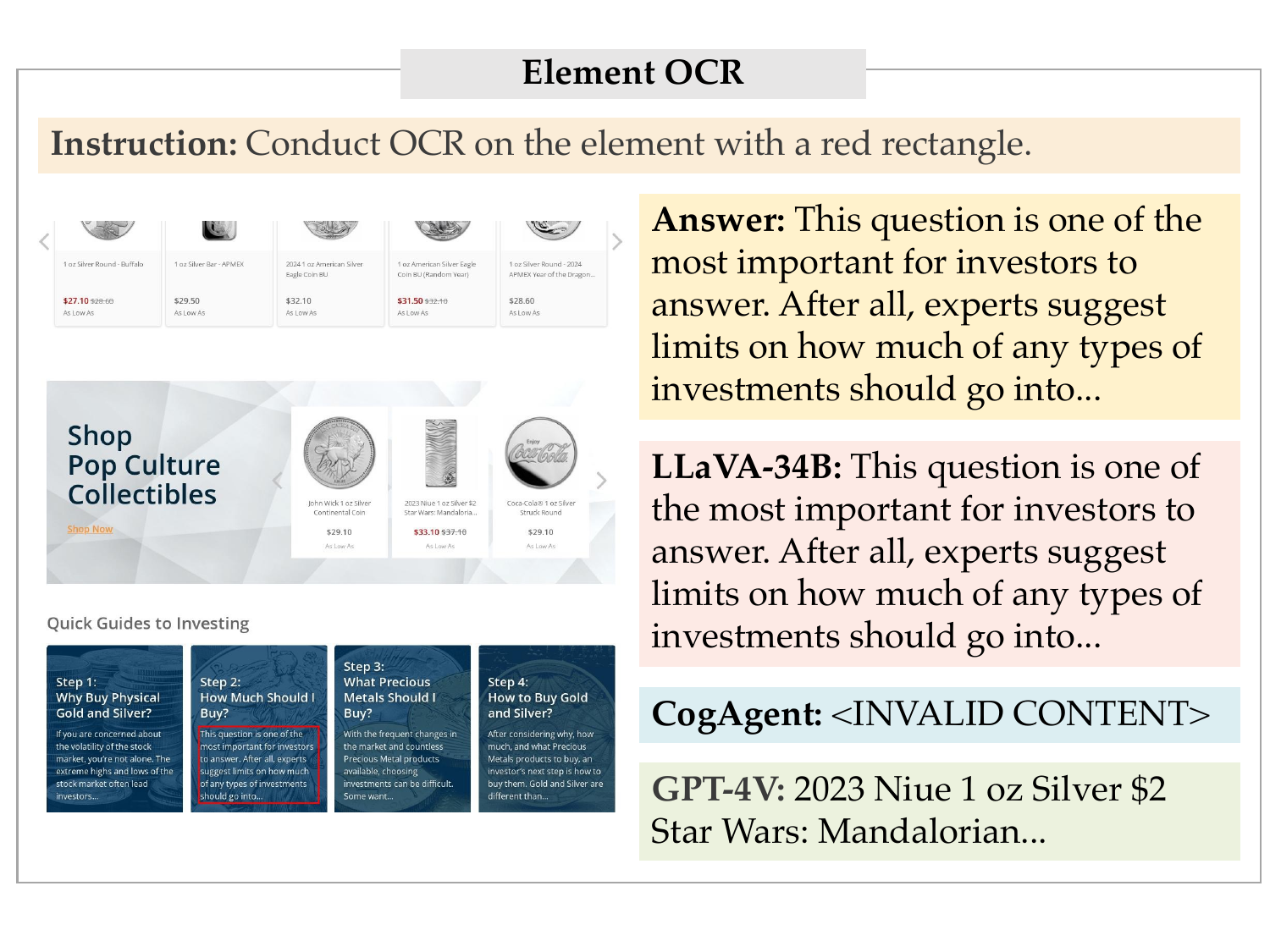}
    \caption{Case study of Element OCR. }
    \label{fig:case_element_ocr}
\end{figure}

Case studies for Captioning, Heading OCR, Element Grounding, and Element OCR are shown in Figure~\ref{fig:case_caption}, Figure~\ref{fig:case_heading_ocr}, Figure~\ref{fig:case_element_grounding}, and Figure~\ref{fig:case_element_ocr}. 

\end{document}

%% file: 1-intro-v2.tex
\section{Introduction}
The web is an indispensable platform for information exchange and interaction, presenting unique challenges and opportunities for multimodal learning. While web content has been a primary source of training data for multimodal large language models (MLLMs)~\citep{2023GPT4VisionSC,team2023gemini,liu2024visual}, a largely overlooked aspect is understanding of websites themselves. Every website is designed to be visually rendered for consumption by human users, with structured layouts, rich textual information, and diverse interactive elements. Enabling MLLMs to accurately comprehend websites would unlock numerous applications in the web domain.

However, evaluating the performance of multimodal models in the web domain is a challenging task. Unlike object- or scene-centric images in most existing benchmarks~\citep{young2014image,goyal2017making,lin2014microsoft,singh2019towards,li2023seed,liu2023mmbench,yu2023mm,yue2023mmmu}, web pages present a complex interplay of visual and textual information, along with interactive elements, requiring models to possess rigorous understanding abilities over hierarchical structures and contextual relationships. Moreover, web elements are often small, numerous, and scattered across the page, demanding fine-grained recognition and accurate spatial reasoning and grounding. The vast diversity of website designs, layouts, and content across different domains further complicates the creation of representative and robust evaluation benchmarks, necessitating the inclusion of a wide range of website categories to ensure the generalizability of the evaluated models.

Despite the growing importance of the web domain in multimodal learning, existing benchmarks fall short of comprehensively evaluating the fundamental capabilities of models in this context. General MLLM benchmarks~\citep{young2014image,liu2023mmbench,yue2023mmmu}, do not adequately capture the unique characteristics of the web domain. On the other hand, web-agent benchmarks, like WebShop~\citep{yao2022webshop}, Mind2Web~\citep{deng2024mind2web}, and (Visual)WebArena~\citep{zhou2023webarena,koh2024visualwebarena}, focus on end-to-end abilities without offering a fine-grained assessment of essential skills such as OCR, semantic understanding, and grounding. Measuring these fine-grained abilities is crucial, as they serve as building blocks for complex web-related tasks, enable targeted improvements, and provide a clearer picture of a model's performance. The lack of granularity in existing benchmarks hinders the development of more capable multimodal models for the web domain, emphasizing the need for a comprehensive evaluation benchmark.

\begin{figure}
  \centering
  \includegraphics[width=\textwidth]{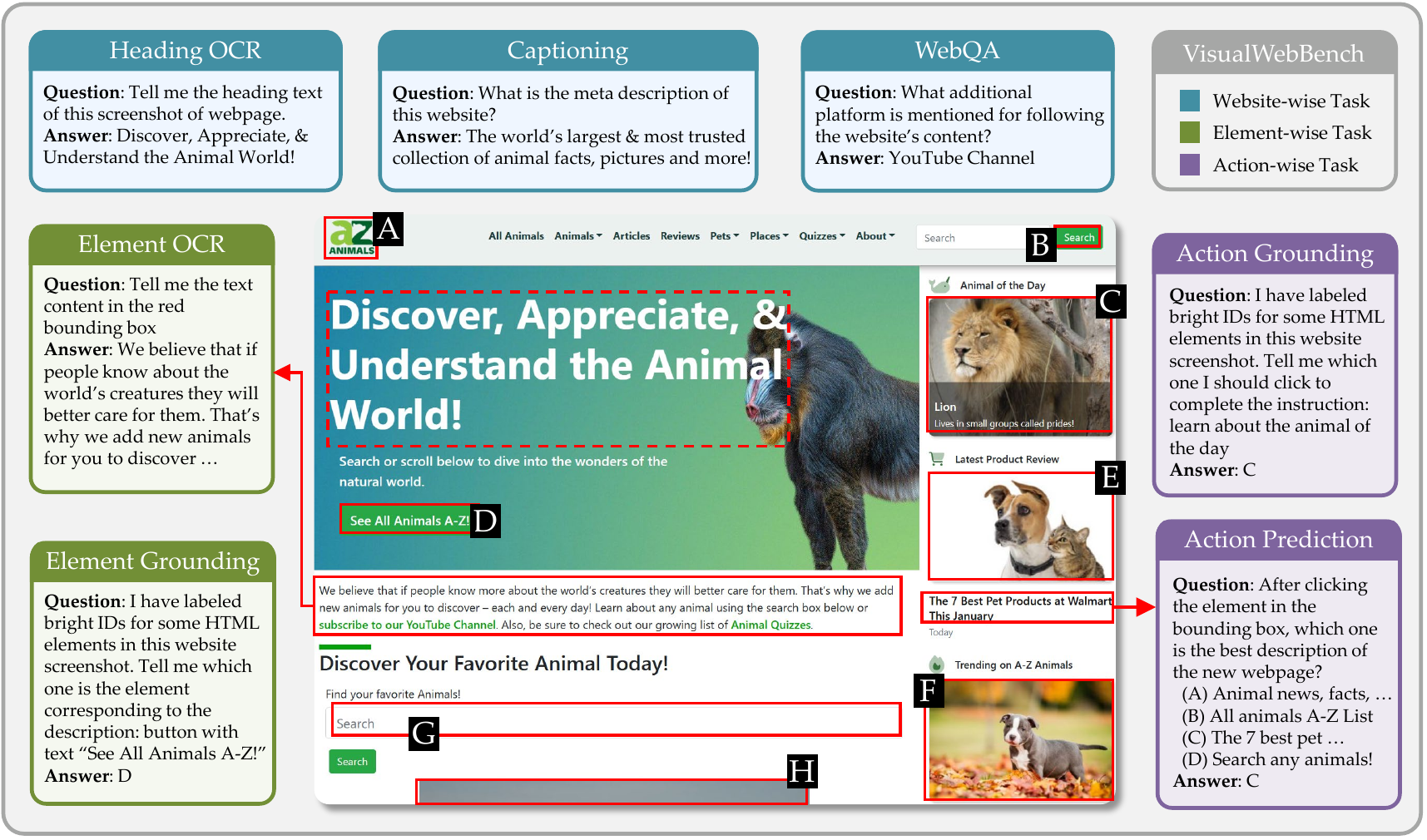}
  \caption{\bench{} contains seven QA-style tasks, covering website, element, action-level understanding, reasoning, and grounding capabilities. 
  }
  \vspace{-15pt}
  \label{fig:main}
\end{figure}

To address these limitations, we introduce \bench{}, a comprehensive multimodal benchmark designed to assess the capabilities of MLLMs in the web domain. Inspired by the human interaction process with web browsers, \bench{} consists of seven tasks that map to core abilities required for web tasks: captioning, webpage QA, heading OCR, element OCR, element grounding, action prediction, and action grounding, as detailed in Figure~\ref{fig:main}. The benchmark comprises 1.5K instances, all uniformly formulated in the QA style, making it easy to evaluate and compare the performance of different MLLMs.

We evaluate 14 open-source MLLMs, Gemini Pro~\citep{team2023gemini}, Claude Sonnet, Claude Opus~\citep{claude}, and GPT-4V(ision)~\citep{2023GPT4VisionSC} on \bench{}; our key findings are as follows:

\begin{itemize}[leftmargin=*, nolistsep]
\setlength{\itemsep}{1mm}
\item \bench{} presents significant challenges for current MLLMs, with GPT-4V and Claude Sonnet achieving average scores of 64.6 and 65.8, respectively, indicating substantial room for improvement.
\item A notable performance gap exists between open-source MLLMs and proprietary counterparts such as GPT-4V and Claude series, with the leading open-source model, LLaVA-1.6-34B, achieving an average score of 50.5.
\item MLLMs' abilities in general domains, such as general reasoning on MMMU~\citep{yue2023mmmu}, and web agent tasks, such as Mind2Web~\citep{deng2024mind2web}, do not correlate much with their performance on \bench{}, highlighting the importance of web-specific benchmarks like \bench{}.
\item The limited image resolution handling capabilities of most open-source MLLMs restrict their utility in web scenarios, where rich text and elements are prevalent.
\item Grounding ability, a crucial skill for developing MLLM-based web applications like autonomous web agents, is a weakness for most MLLMs.
\end{itemize}


In summary, \bench{} offers a standardized benchmark for evaluating MLLMs in web understanding, enabling the development of more capable and efficient models, autonomous web agents, and web-related applications.

%% file: 4-related.tex
\section{Related Work}

\begin{figure}
  \centering
  \includegraphics[width=\textwidth]{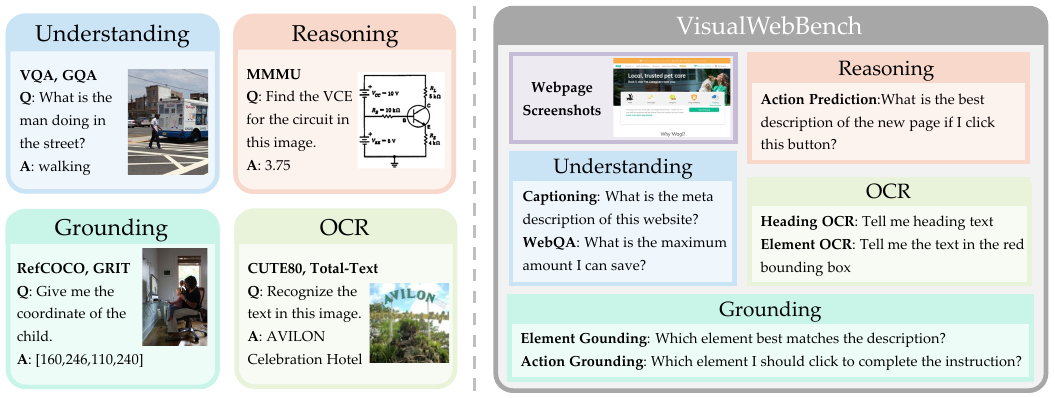}
  \caption{Comparison between \bench{} (right) and other multimodal benchmarks (left).
  }
  \label{fig:compare}
\end{figure}

Before detailing \bench{}, we briefly outline its differences with existing MLLM benchmarks, also outlined in Table~\ref{fig:compare}.

\subsection{MLLM Benchmarks}

In concert with improvements in these MLLMs, benchmarks have also evolved.
These range from traditional single task benchmarks like VQA~\citep{antol2015vqa,goyal2017making}, RefCOCO~\citep{mao2016generation}, and Flickr30K~\citep{young2014image}, to more holistic evaluation benchmarks like LAMM~\citep{yin2024lamm}, MMBench~\citep{liu2023mmbench}, and MMMU~\citep{yue2023mmmu}, recently.
In this work, we focus on images in web-based scenarios characterized by structured layouts, copious textual data, and diverse interactive elements, which pose new challenges for current MLLMs.
The most closely related scenario to this work is GUI-based tasks, exemplified by Screen2Words~\citep{wang2021screen2words}, Widget Captioning~\citep{li2020widget}, and WebSRC~\citep{chen2021websrc} which is a web-based VQA dataset.
Different from previous works, \bench{} offers a comprehensive evaluation for MLLMs, spanning perception, comprehension, grounding, and reasoning capabilities.

\subsection{Web Agent Benchmarks}

As a vital aspect of daily life, methods that perform various tasks in web scenarios have garnered widespread attention from researchers.
Earlier efforts introduce simplified simulated environments for web navigation tasks, such as MiniWob++~\citep{liu2018reinforcement} and WebShop~\citep{yao2022webshop}.
Recently, Mind2Web~\citep{deng2024mind2web}, WebArena~\citep{zhou2023webarena}, VisualWebArena~\citep{koh2024visualwebarena} construct realistic and reproducible web environments to facilitate the development of web agents.
There are also various studies to improve the web understanding or grounding capabilities of MLLMs~\citep{gao2024enhancing,kil2024dual} or develop agents for autonomous web navigation~\citep{hong2023cogagent,zheng2024gpt,cheng2024seeclick,he2024webvoyager}.
Despite their success, the community still lacks a comprehensive evaluation of MLLMs' basic performance in web scenarios, including perception, understanding, grounding, and reasoning.

%% file: 2-bench.tex
\section{The \bench{} Benchmark}

\subsection{Overview of \bench{}}


\begin{figure}[t]
\centering
    \begin{minipage}{0.32\textwidth}
    \centering
    \includegraphics[width=0.98\textwidth]{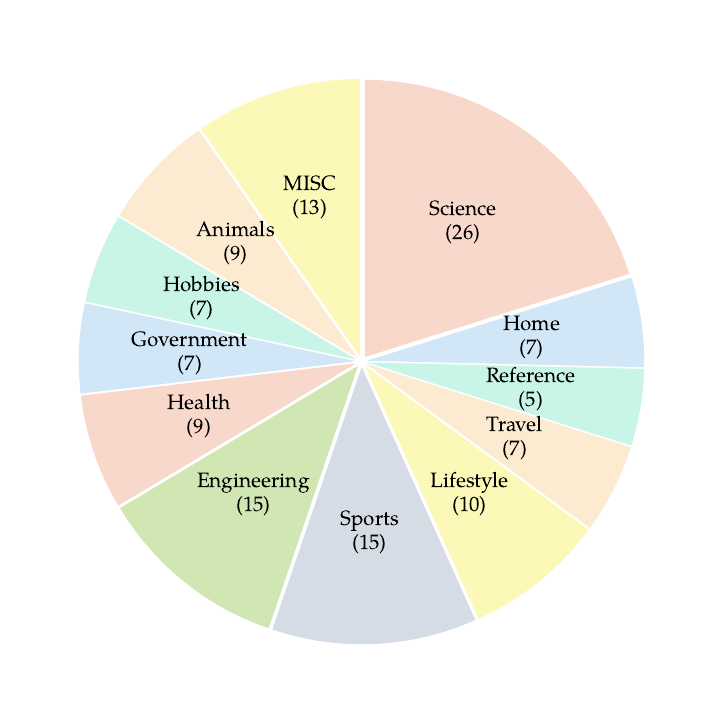}
    \end{minipage}
    \hfill
    \begin{minipage}{0.65\textwidth}
    \centering
    \resizebox{\textwidth}{!}{
    \begin{tabular}{lllll}
    \toprule
    \textbf{Task} & \textbf{Level} & \textbf{Capability} & \textbf{Metric} & \textbf{\#Num.} \\
    \midrule
     Captioning & \multirow{3}{*}{Webpage} & Understanding & ROUGE-L & 134 \\
    WebQA & & Understanding & F1 & 314 \\
    Heading OCR & & OCR & ROUGE-L & 46 \\
    \midrule
    Element OCR & \multirow{2}{*}{Element} & OCR & ROUGE-L & 245 \\
    Element Grounding & & Grounding & Accuracy & 413 \\
    \midrule
    Action Prediction & \multirow{2}{*}{Action} & Reasoning & Accuracy & 281 \\
    Action Grounding & & Grounding & Accuracy & 101 \\
    \midrule
    Total & - & - & - & 1534 \\
    \bottomrule
    \end{tabular}
    }
    \end{minipage}
\centering
\caption{Overview of \bench{}. Left: Domain distribution. The numbers represent the count of sub-domains within each domain. Right: Tasks in \bench{}. In the evaluated capabilities, U, R, G means understanding, reasoning, and grounding, respectively. }
\label{fig:overview}
\end{figure}

We present \bench{}: a multimodal benchmark designed to thoroughly evaluate the understanding and grounding capabilities of MLLMs in web scenarios. The proposed \bench{} possesses the following features:
1) \textbf{Comprehensiveness}: \bench{} spans 139 websites with 1.5K samples, encompassing 12 different domains (e.g., travel, sports, hobby, lifestyle, animals, science, etc.) and 87 sub-domains.
2) \textbf{Multi-granularity}: \bench{} assesses MLLMs at three levels: website-level, element-level, and action-level.
3) \textbf{Multi-tasks}: \bench{} encompasses seven tasks designed to evaluate the understanding, OCR, grounding, and reasoning capabilities of MLLMs.
4) \textbf{High quality}: Quality is ensured through careful human verification and curation efforts.
The domain distribution and statistics of our benchmark are presented in Figure~\ref{fig:overview}.

\subsection{Website Selection}


To ensure comprehensive coverage across diverse domains and top-ranking websites, our website selection process is conducted based on SimilarWeb\footnote{\href{https://www.similarweb.com}{https://www.similarweb.com}}.
We start from 12 top-level domains in SimilarWeb such as Science, Engineering, Sports, Lifestyle, and more, which are subsequently broken down into 87 sub-domains.
Then we manually select representative websites from the top-5 most ranking websites in each sub-domain.
Our selection criteria prioritize websites with rich interactive elements, including images and buttons, while excluding those that have been used in prior web agent benchmarks like Mind2Web and WebArena.
We use Playwright\footnote{\href{https://github.com/microsoft/playwright}{https://github.com/microsoft/playwright}} to render and save the websites automatically.


\subsection{Task Construction}
\label{sec:task-construction}

This section details the proposed seven tasks of \bench{} and the process of constructing data for each task; examples are shown in \autoref{fig:main}.

\noindent\textbf{Captioning.} 
To evaluate the MLLMs' ability to comprehend and summarize the content of a webpage screenshot, we propose a webpage captioning task.
The meta description, i.e., \verb|<meta name="description">| tags in the head section of HTML, is a brief snippet of text that helps humans or search engines understand the content of websites. However, the quality of extracted meta descriptions cannot be ensured and their styles are pretty different on diverse websites. For example, some meta descriptions only consist of a list of keywords or a short title of the website, instead of a natural language description. Hence, we instruct GPT-4V to generate a better meta description in a unified style as the caption, given both the screenshot and the extracted meta description. The final captions are verified and curated by the authors.

\noindent\textbf{WebQA.} 
To assess the understanding capabilities of MLLMs in the web scenario, \bench{} involves a webpage QA task, where the MLLM will answer open questions that demand a thorough comprehension of the visual layout.
Human annotators are instructed to examine each screenshot and craft up to five challenging questions which satisfy: 1) A degree of reasoning ability is required to answer the question, 2) The answers should be precise and objective.


\noindent\textbf{Heading OCR.} 
This task requires MLLMs to locate and recognize the text of the heading of a website.
Different from the traditional OCR task where a target element is given, as shown in Figure~\ref{fig:main}, the input of heading OCR is simply a raw screenshot, and the expected output is the heading content.
The ground-truth target is automatically extracted from the first \verb|<h1>| element in the HTML.

\noindent\textbf{Element OCR.} 
This task evaluates the capability of MLLMs to conduct OCR on lengthy texts. 
Firstly, we traverse the HTML DOM tree and extract the bounding boxes and textual description of each element
Then, we select elements whose text descriptions comprise over 20 words. 
The task input consists of a screenshot with a bounding box indicating the position of the element to be recognized. 

\noindent\textbf{Element Grounding.}
Grounding, or Referring Expression Comprehension (REC), is a crucial image-text alignment capability, particularly for MLLMs interacting with web environments.
Given a description of an HTML element, MLLM needs to locate the corresponding region in the screenshot.
However, our preliminary studies reveal that current MLLMs struggle to directly give the coordinate of the target's bounding box (see ~\ref{sec:ground_analysis}).
Inspired by \citet{yang2023set}, we adopt a simplified setting where eight candidate bounding boxes are presented. Differently, the candidate elements here are extracted automatically using Playwright, with each assigned an alphanumeric ID.
MLLMs are then prompted to select the box that best matches the given element description.
The element description, golden bounding box, and negative bounding boxes of randomly chosen elements are automatically extracted from the webpage.


\noindent\textbf{Action Prediction.} 
This task asks MLLMs to predict the title of the redirected website after clicking an element, in a multi-choice QA way. 
In terms of the construction process, firstly, we employ Playwright to click all clickable elements within the web page and save the \verb|<title>| or \verb|<meta name="title">| tag as titles of new redirected web pages. 
Subsequently, we randomly sample seven additional elements distinct from the target element and take the titles of their respective redirect destinations as negative choices. 
Cases where a click does not lead to a title change are omitted from consideration.
The task presents input in the form of screenshots highlighting the clickable target with a red bounding box. 
Accompanying each screenshot is eight choices, each labeled with a letter. 
The ground truth output is the letter corresponding to the correct answer.

\noindent\textbf{Action Grounding.}
In addition to directly grounding elements from their descriptions, we further introduce the action grounding task.
In this task, the MLLMs are given a human instruction, such as ``search for the hotels in NYC'', and are prompted to determine the correct element to click to fulfill the instruction.
Similar with \textbf{Element Grounding}, MLLMs take in a screenshot containing bounding boxes of eight candidate elements and select the most appropriate one.
The task data is completed by seven experienced annotators, and an annotation tool is developed to streamline the annotation workflow. 
Further details about the annotation tool and the annotation process can be found in Appendix~\ref{sec:annotation_tool}.



All tasks above adopt a VQA-style formulation similar to customary multimodal benchmarks.
All screenshots in \bench{} are unified in a standard 1280-pixel width.
All samples of our benchmark undergo careful verification and curation through a collaborative effort and a division of tasks by two authors. See Appendix \ref{sec:curation} for more details.

\subsection{Evaluation Metrics}

We adopt different evaluation metrics for different tasks in \bench{}.
For open-ended generation tasks, ROUGE-L~\citep{lin2004rouge} is used to measure the quality of the generated responses. 
For the WebQA task, SQuAD style F1~\citep{rajpurkar2016squad} is employed as the evaluation metric.
For multiple-choice tasks, we measure accuracy.

%% file: 3-experiment.tex
\section{Experiments}

\begin{table}[t]
\centering
\small
\resizebox{\textwidth}{!}{
\begin{tabular}{lccccccccc}
\toprule
\multirow{2}{*}{\textbf{Model}} & \multicolumn{3}{c}{\textbf{Website}} & \multicolumn{2}{c}{\textbf{Element}} & \multicolumn{2}{c}{\textbf{Action}} & \multirow{2}{*}{\textbf{Average}} \\
\cmidrule(l){2-4} \cmidrule(l){5-6} \cmidrule(l){7-8}
& Caption & WebQA & HeadOCR & OCR & Ground & Prediction & Ground &  \\
\midrule
\multicolumn{9}{c}{\textbf{General MLLMs}} \\

\midrule
Otter            & 5.3           & 0.7           & 3.5           & 0.5           & 0.7           & 14.6          & 0.0           & 3.6           \\
InstructBLIP-13B & 11.6          & 5.2           & 7.6           & 6.0           & 11.4          & 11.4          & 17.5          & 10.1          \\
BLIP-2           & 11.0          & 5.2           & 20.6          & 2.6           & 15.5          & 14.9          & 8.7           & 11.2          \\
Fuyu-8B          & 3.5           & 5.2           & 5.8           & 12.4          & 19.4          & 13.2          & 15.5          & 10.7          \\
Yi-VL-6B         & 8.0           & 14.3          & 43.8          & 3.5           & 16.2          & 13.9          & 13.6          & 16.2          \\
LLaVA-1.5-7B     & 15.3          & 13.2          & 41.0          & 5.7           & 12.1          & 17.8          & 13.6          & 17.0          \\
mPLUG-Owl2       & 12.7          & 19.9          & 51.6          & 7.2           & 11.9          & 23.1          & 3.9           & 18.6          \\
LLaVA-1.5-13B    & 20.0          & 16.2          & 41.1          & 11.8          & 15.0          & 22.8          & 8.7           & 19.4          \\
SPHINX           & 13.7          & 11.6          & 48.1          & 7.7           & 18.4          & 14.2          & 22.3          & 19.4          \\
Qwen-VL          & 21.8          & 32.2          & 48.4          & 13.4          & 14.0          & 26.7          & 10.7          & 23.9          \\
CogVLM           & 16.6          & 30.6          & 65.9          & 10.0          & 17.7          & 11.7          & 23.3          & 25.1          \\
VILA-13B         & 12.7          & 28.8          & 67.9          & 12.6          & 16.5          & 36.3          & 16.5          & 27.3          \\
DeepSeek-VL-7B   & 18.1          & 30.0          & 63.4          & 18.1          & 16.2          & 35.2          & 15.5          & 28.1          \\
LLaVA-1.6-7B     & 27.0    & 39.8          & 57.3          & 54.8          & 31.7          & 30.6          & 10.7          & 36.0          \\
LLaVA-1.6-13B    & 26.5          & 44.5          & 52.8          & 56.1          & 31.7          & 48.4          & 15.5          & 39.4          \\
LLaVA-1.6-34B    & 24.3          & 48.2          & 67.1          & 71.9    & 43.1          & \textbf{74.0}    & 25.2          & 50.5          \\
\addlinespace[0.1em]\hdashline\addlinespace[0.3em]
Gemini Pro       & 25.0          & 55.5    & \textbf{75.1} & 65.4 & 44.3    & 26.7          & 43.7    & 48.0    \\
Claude Sonnet & \underline{28.9} & \textbf{81.8} & \underline{70.3} & \textbf{89.2} & \textbf{68.8} & 63.4 & \underline{58.3} & \textbf{65.8} \\
Claude Opus & 26.7 & \underline{75.4} & 63.7 & \underline{87.1} & 57.7 & 60.4 & 38.8 & 58.5 \\
GPT-4V(ision)    & \textbf{34.5} & 75.0 & 68.8    & 62.8          & \underline{67.5} & \underline{67.6} & \textbf{75.7} & \underline{64.6} \\
\midrule
\multicolumn{10}{c}{\textbf{GUI Agent MLLMs}} \\
\midrule
SeeClick         & 0.0           & 19.6          & 34.8          & 0.0           & 9.9           & 1.8           & 1.9           & 9.7           \\
CogAgent-Chat    & 16.3          & 53.3          & 20.2          & 32.4          & 41.6          & 13.5          & 23.3          & 28.7          \\

\bottomrule
\end{tabular}
}
\caption{Overall results of different models on \bench{} benchmark. The best-performing model is \textbf{in-bold}, and the second best is \underline{underlined}. The maximum of the metrics is 100.}
\label{tab:main_result}
\end{table}
\subsection{Evaluated MLLMs}
We evaluate 14 open-source general MLLMs on \bench{} (See Appendix~\ref{sec:experiment-lmms} for model details).
By default, for each model family, we use the largest available checkpoint.
We consider three scales of LLaVA, 7B, 13B, and 34B, for model scaling analysis.
Several strong close-source MLLMs, Gemini Pro~\citep{team2023gemini}, Claude series~\citep{claude}, and GPT-4V(ision)~\citep{2023GPT4VisionSC}, are also included for evaluation.

Recent studies have introduced several MLLMs tailored to create agents for GUI tasks, such as web and smartphones~\citep{cheng2024seeclick,hong2023cogagent,gao2024enhancing}.
Therefore, we consider two open-source GUI-specialized MLLMs for evaluation:
SeeClick~\citep{cheng2024seeclick} is developed by GUI grounding pretraining based on Qwen-VL~\citep{bai2023qwen}.
CogAgent~\citep{hong2023cogagent} is built upon CogVLM~\citep{wang2023cogvlm}, focusing on GUI interpretation and navigation, with support for high-resolution image inputs.

\subsection{Main Results}
\label{sec:main_exp}
In this section, we present a comprehensive comparison of different MLLMs on \bench{} in Table~\ref{tab:main_result}.
From the results, we highlight the following findings.

\textbf{Challenging Nature of Web Tasks:}
Even the most powerful MLLM, GPT-4V, achieves an average score of only 64.6 on \bench{}, leaving ample room for improvement.
For tasks requiring strong reasoning and grounding abilities (Action Prediction and Action Grounding), many MLLMs struggle to surpass random chance (12.5).
This underscores that current models cannot effectively handle many tasks within the web scenario.

\textbf{Disparity between Open-source and Proprietary MLLMs:}
GPT-4V and Claude outperform open-source MLLMs including GUI agent MLLMs by a large margin, highlighting a discernible gap in the capabilities of current open-source MLLMs compared to proprietary ones like GPT-4V. 
Meanwhile, LLaVA-1.6-34B achieves a commendable result (50.5) and beats all other open-source MLLMs, even outperforming the performance of Gemini Pro (48.0).
Notably, we find Claude Sonnet surpasses Opus on all tasks in \bench{}, suggesting that Sonnet may possess more powerful capabilities in web scenarios.

\textbf{Scaling Leads to Better Performance:}
Compared with the 7B and 13B versions of LLaVA-1.6, the 34B model achieves a performance boost across almost all tasks, reaching an average score of 50.5. 
Although there are factors other than scale, such as different backbone LLMs, this indicates that increasing model size is a promising avenue for enhancing the capabilities of open-source MLLMs in web-related tasks.


\begin{figure}[t]
    \centering
    \begin{minipage}[b]{0.47\linewidth} 
        \includegraphics[width=6.0cm]{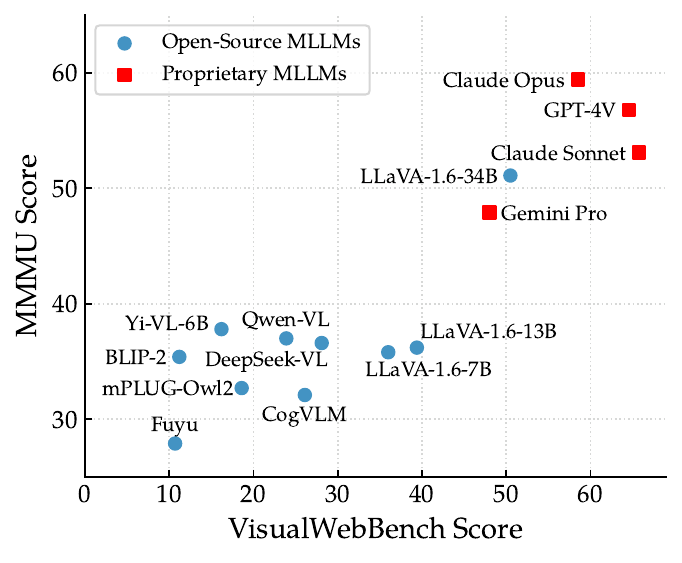}
        \caption{Scores of MLLMs on \bench{} and MMMU.}
        \label{fig:analysis_MMMU_webbench}
    \end{minipage}
    \hfill 
    \begin{minipage}[b]{0.47\linewidth} 
        \includegraphics[width=6.0cm]{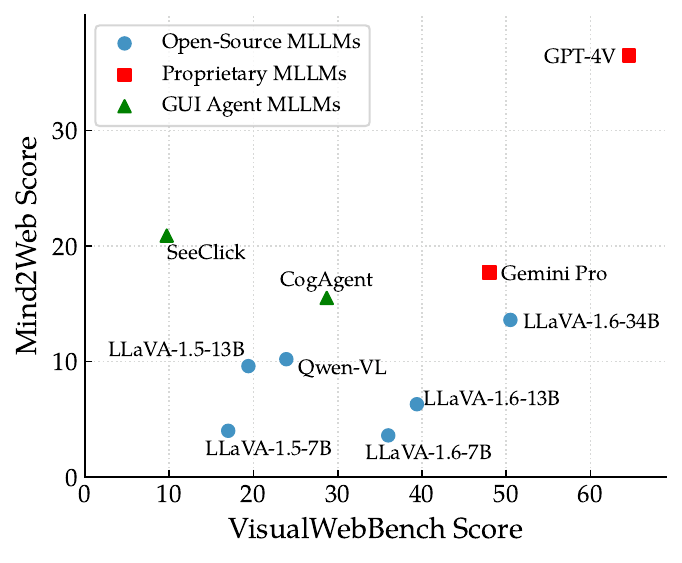}
        \caption{Scores of MLLMs on \bench{} and Mind2Web. }
        \label{fig:mind2web}
    \end{minipage}
\end{figure}


\textbf{General MLLMs vs. GUI Agent MLLMs:}
\label{sec:general_vs_gui}
SeeClick and CogAgent are two MLLMs pre-trained on GUI grounding tasks.
However, we observe that these GUI agent MLLMs do not exhibit significant performance improvement.
For example, SeeClick fails to outperform Qwen-VL, its base MLLM, across all tasks.
Notably, we find these models suffer catastrophic forgetting~\citep{wang2024comprehensive} on general instruction following capability after training on GUI grounding data.
These results underscore the necessity for more general GUI-specific training techniques to enhance the MLLMs' performance in the web scenario.
To further investigate the effectiveness of GUI grounding training, we perform a comprehensive comparison of various grounding settings in Section~\ref{sec:ground_analysis}.


\subsection{Correlations with General Scenario and Agent Benchmarks}
We delve into the relationship between the performance of MLLMs in the web scenario and that in general and agent scenarios.
Specifically, we use MMMU~\citep{yue2023mmmu} as the proxy of MLLMs' capability in general scenario\footnote{The overall score on the validation set of MMMU is used for comparison.}, and Mind2Web~\citep{deng2024mind2web} is used for the evaluation of the agent scenario. 

While Figure~\ref{fig:analysis_MMMU_webbench} somewhat suggests some correlation between VisualWebBench score and MMMU score, the relationship does not appear to be significant. 
In other words, performing well in the general domain does not necessarily guarantee the same trend in the web scenario. 
For example, while Yi-VL-6B and BLIP2 outperform CogVLM on MMMU, they fall short in achieving a good score on \bench{}.
It is also noteworthy that LLaVA-1.6-34B performs well on both tasks, nearly matching the performance level of GPT-4V.

As illustrated in Figure~\ref{fig:mind2web}, generally, VisualWebBench scores are higher than those of Mind2Web\footnote{Detailed experimental results are included in Appendix~\ref{sec:webagent}.}, which demonstrates that there still exists large room for improvement of agent ability empowered by webpage understanding, grounding abilities, as well as other abilities like planning. GUI agent MLLMs tend to exhibit overfitting in terms of agent capability, resulting in underperformance in understanding web pages.

\begin{wrapfigure}{r}{5.5cm}
    \centering
    \includegraphics[width=0.95\linewidth]{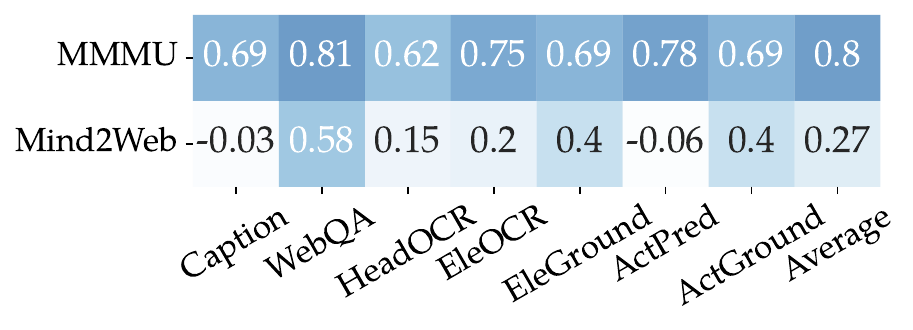}
    \caption{Correlations between \bench{} and MMMU, Mind2Web, respectively.}
    \vspace{-10pt}
    \label{fig:correlation_mmmu_mind2web}
\end{wrapfigure}

Moreover, in Figure \ref{fig:correlation_mmmu_mind2web}, we conduct an in-depth correlation analysis between MMMU and \bench{} seven subtasks, as well as a similar analysis between Mind2Web and \bench{} tasks. For MMMU, the correlations are generally strong. Specifically, the two subtasks requiring heavy reasoning, WebQA and Action Prediction, strongly correlate with MMMU.
For Mind2Web, the correlation between scores on \bench{} and those on Mind2Web is low, even exhibiting two negative correlations in Captioning and Action Grounding. 
These findings suggest that \bench{} offers a different evaluation perspective for MLLMs in the web scenario.  



\subsection{Correlation Between \bench{} Tasks}

\begin{wrapfigure}{r}{5.5cm}
    \centering
    \includegraphics[width=0.95\linewidth]{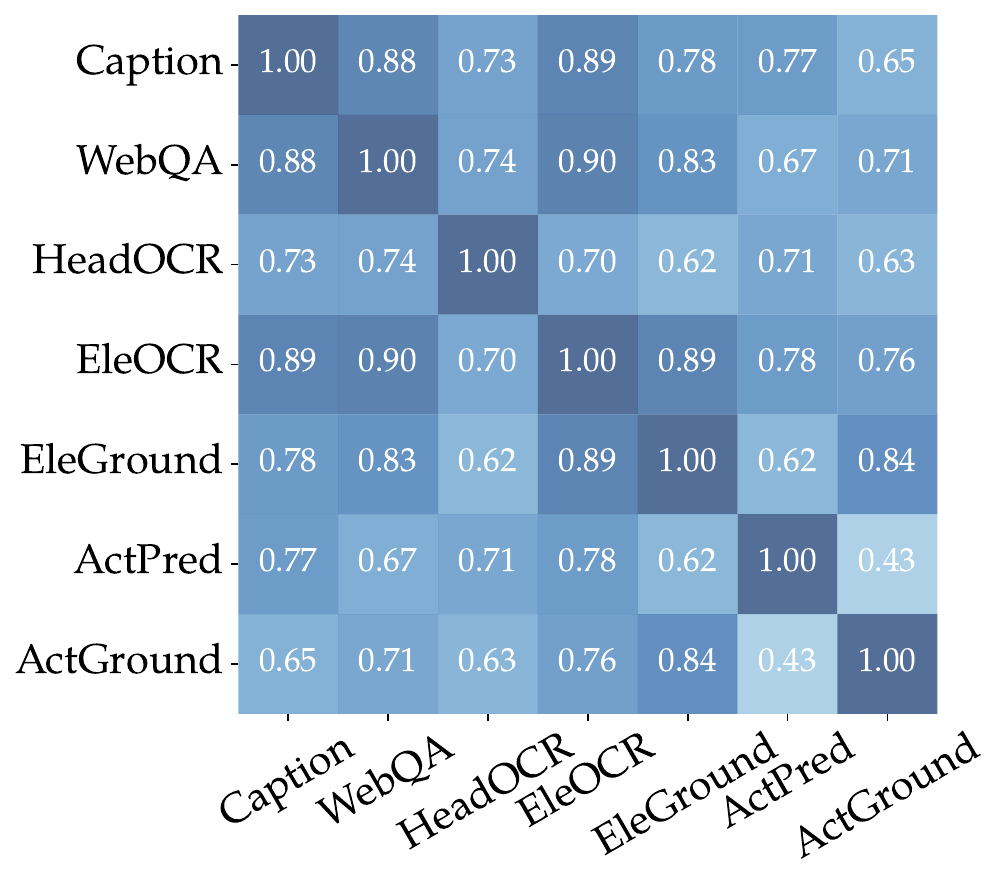}
    \caption{Correlations between 7 subtasks in \bench{}.}
    \vspace{-10pt}
    \label{fig:correlation}
\end{wrapfigure}

Figure~\ref{fig:correlation} illustrates the correlations between tasks in \bench{}.
This analysis reveals a strong correlation among specific tasks, namely Captioning, WebQA, and Element OCR, all demanding a comprehensive understanding of textual content within webpages.
In contrast, Action Prediction and Action Grounding tasks exhibit a minimal correlation, implying distinct skill sets necessary for predicting action outcomes versus pinpointing elements for actions.
Moreover, Action Grounding seems to be less correlated with all other non-grounding tasks, highlighting its distinctive and specialized skill requirements.


\subsection{Analysis of Image Resolution}

\begin{figure}[t]
    \centering
    \includegraphics[width=0.9\textwidth]{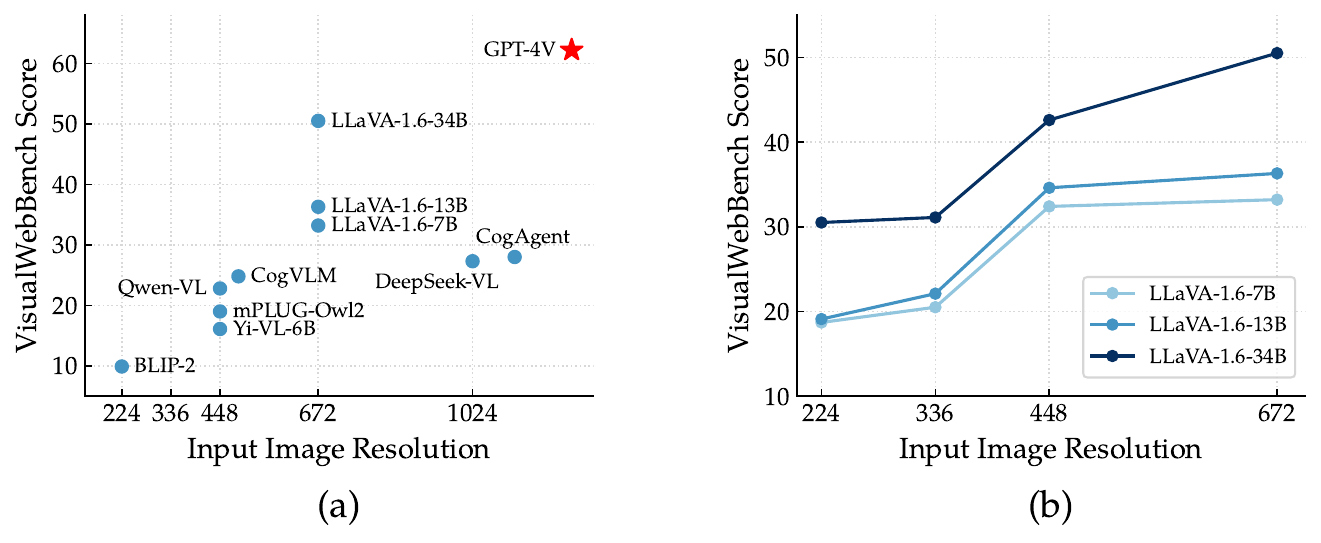}
    \caption{The effect of image resolution on VisualWebBench score. The input image resolution of GPT-4V is unknown.}
    \label{fig:analysis_resolution}
\end{figure}

Most current MLLMs can only process low-resolution images, typically 448$\times$448.
However, the screenshots in \bench{} are captured in high resolution (1280 pixels in width), presenting challenges in identifying intricate details at lower resolutions.
In this section, we explore the effects of input resolution on model performance.
We plot the relation of max input image resolution and \bench{} scores for different MLLMs in Figure~\ref{fig:analysis_resolution}(a).
Notably, MLLMs with higher input resolution generally achieve higher scores.
For instance, DeepSeek-VL with 1024$\times$1024 resolution achieves a higher score than Qwen-VL with 448$\times$448 resolution.

Based on LLaVA-1.6 series models, we further conduct a formal ablation study on input image resolution.
As depicted in Figure~\ref{fig:analysis_resolution}(b), a significant performance improvement is observed as input image resolutions increase for all three model sizes.
Additionally, the models exhibit greater benefits when increasing resolution from 336 to 448, compared with from 448 to 672.
This finding suggests that, for LLaVA-1.6, a resolution of 448$\times$448 stands as the minimal requirement to achieve adequate performance in web-related tasks.

\subsection{Analysis of Grounding Capability}
\label{sec:ground_analysis}
In our experiments in Section~\ref{sec:main_exp}, for Element and Action Grounding tasks, we provide eight candidate elements and use a multiple-choice setting to evaluate different MLLMs.
However, in many applications, the screenshots of webpages cannot be annotated with candidate bounding boxes.
Hence, we evaluate the grounding capability in unannotated images by framing the grounding tasks as a Referring Expression Comprehension (REC) problem, where the MLLMs must generate the position (bounding box $[x_1,y_1,x_2,y_2]$ or central point coordinate $[x,y]$) of the selected HTML element.
For the setting of the bounding box, we follow the standard REC task and use AP$_{50}$~\citep{lin2014microsoft} as the metric.
For the setting of point prediction, a predicted point is regarded as correct if it falls into the true bounding box.

As Table~\ref{tab:ground} shows, GUI agent MLLMs significantly outperform general MLLMs (e.g., LLaVA-1.6 and GPT-4V) in generating the positions (Bbox or point) of target elements, confirming the efficacy of grounding pre-training through the point or bounding box prediction.
For other MLLMs that have been trained on general grounding data like RefCOCO, they still fail to accurately give the coordinates of the correct elements.

\begin{table}
\centering
\resizebox{0.8\textwidth}{!}{
\begin{tabular}{lcccccc}
\toprule
\multirow{2}{*}{\textbf{Model}} & \multicolumn{3}{c}{\textbf{Element Ground}} & \multicolumn{3}{c}{\textbf{Action Ground}} \\
& Multi-choice & Bbox & Point & Multi-choice & Bbox & Point \\
\midrule
Fuyu-8B       & 19.4 & 0.0  & 0    & 15.5 & 0.0  & 0.0  \\
VILA-13B      & 16.5 & 1.0  & 7.8  & 16.5 & 0.0  & 5.9  \\
LLaVA-1.6-7B  & 31.7 & 0.2  & 4.6  & 10.7 & 0.0  & 5.9  \\
LLaVA-1.6-13B & 31.7 & 0.0  & 0.7  & 15.5 & 1.0  & 5.9  \\
LLaVA-1.6-34B & 43.1 & 1.7  & 10.7 & 25.2 & 3.0  & 10.9 \\
Qwen-VL       & 14.0 & 1.5  & 3.9  & 10.7 & 0.0  & 3.0  \\
GPT-4V(ison)  & 67.5 & 0.2  & 1.5  & 75.7 & 0.0  & 1.0  \\
\midrule
SeeClick      & 9.9  & 0.0  & 70.0 & 1.9  & 0.0  & 42.6 \\
CogAgent-Chat & 41.6 & 29.3 & 46.3 & 23.3 & 36.6 & 58.4 \\
\bottomrule
\end{tabular}
}
\caption{Three evaluation settings for grounding tasks. ``Multi-Choice'' is the default setting in \bench{}. ``Bbox'' and ``Point'' denote the setting of predicting the coordinate of the target bounding box and central point, respectively.
}
\label{tab:ground}
\end{table}

\subsection{Case Studies}

\begin{figure}[t]
    \centering
    \begin{minipage}[b]{0.47\linewidth} 
        \includegraphics[width=1\textwidth]{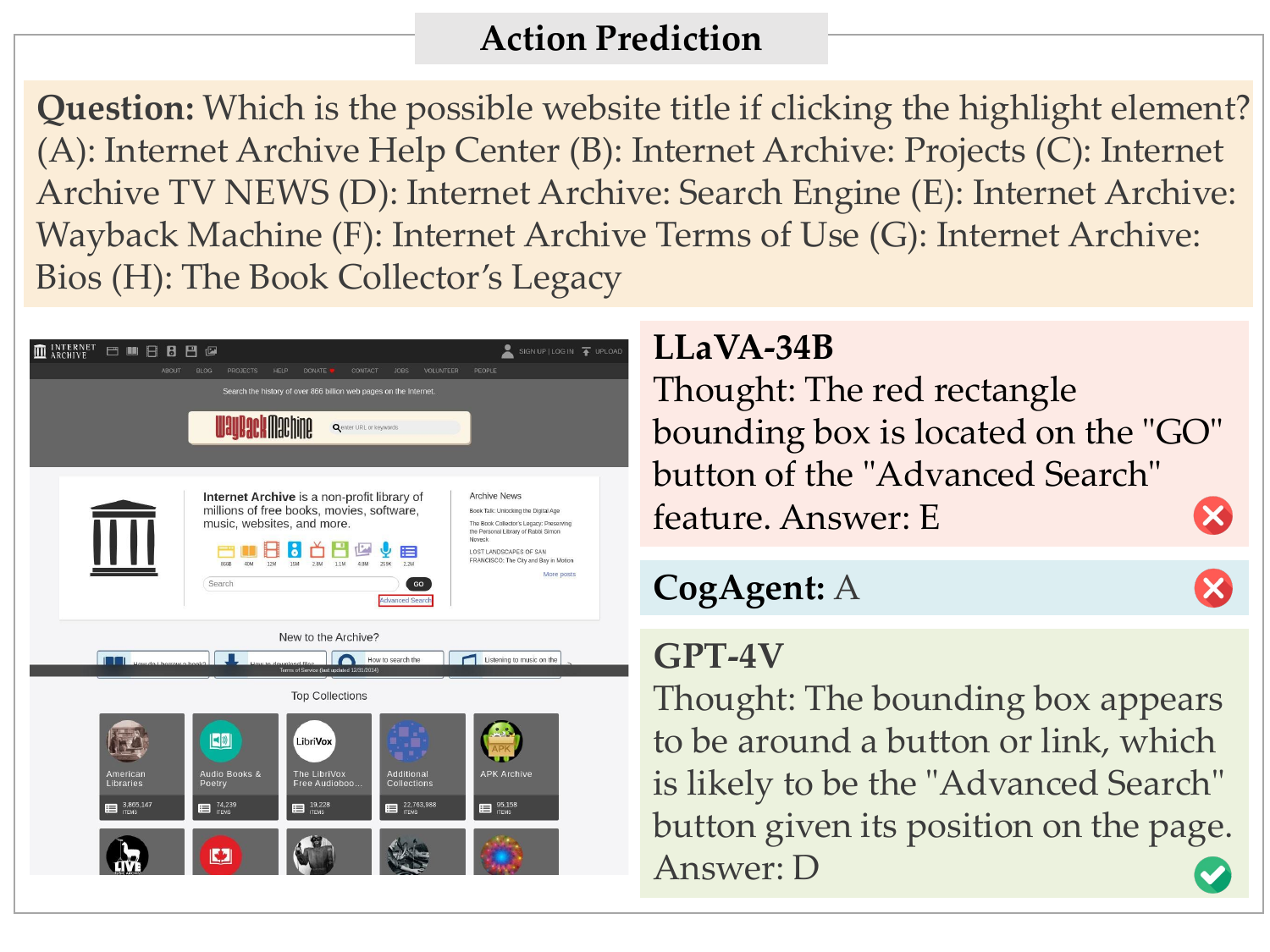}
        \caption{Case study of Action Prediction. }
        \label{fig:case_actpred}
    \end{minipage}
    \hfill 
    \begin{minipage}[b]{0.47\linewidth} 
        \includegraphics[width=1\textwidth]{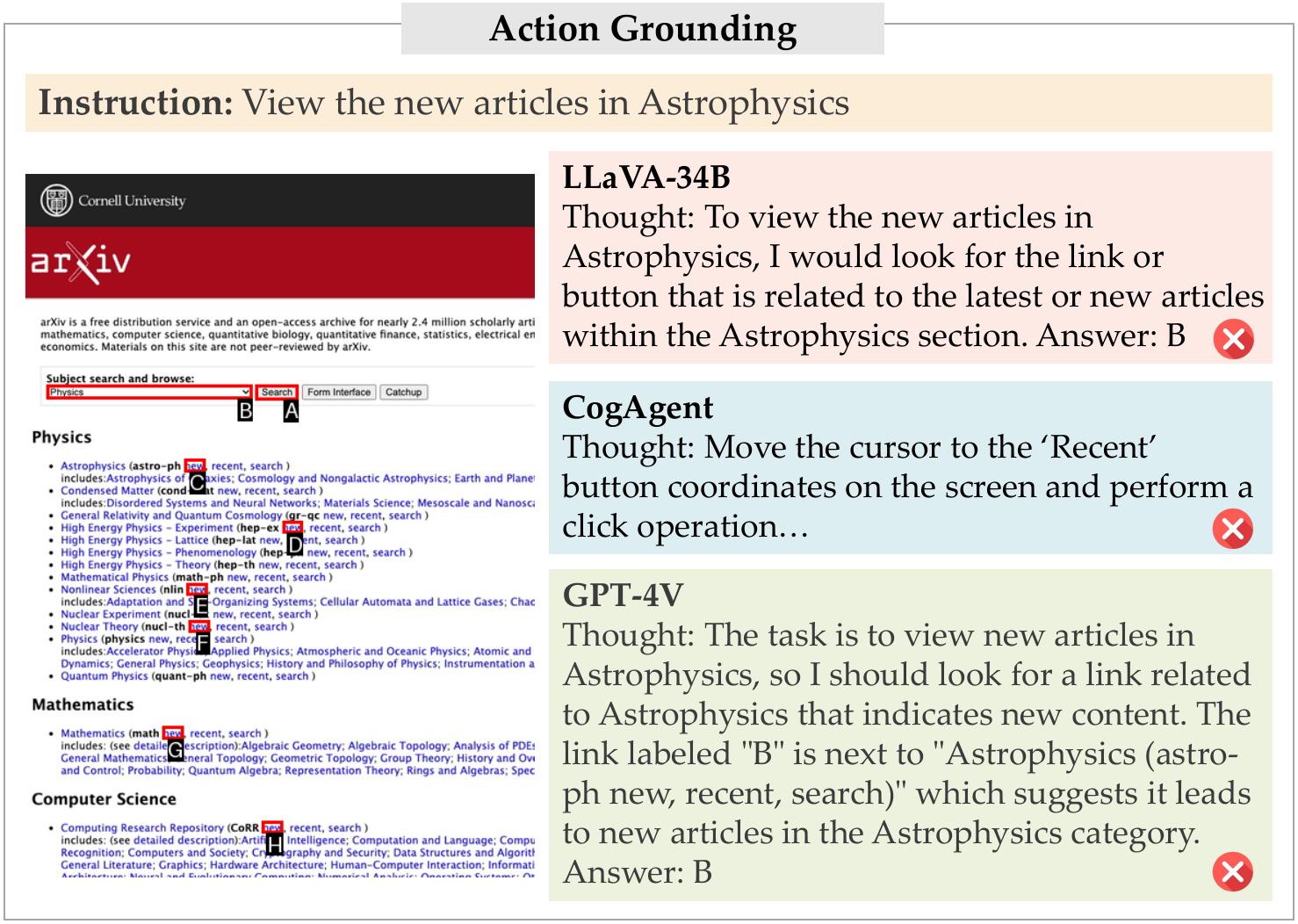}
        \caption{Case study of Action Grounding. }
        \label{fig:case_actground}
    \end{minipage}
\end{figure}


We show a few case studies for LLaVA-1.6-34B, CogAgent, and GPT-4V on action prediction and action grounding tasks. 
For the action prediction task (Figure~\ref{fig:case_actpred}), CogAgent generates a wrong choice without any explanation, while LLaVA locates a wrong element ("Go" button). 
Notably, GPT-4V shows a reasonable thinking process and the correct answer.
For action grounding (Figure~\ref{fig:case_actground}), despite LLaVA and GPT-4V generating reasonable thought processes, all three models fail to answer correctly.
See Appendix~\ref{sec:case_webqa} for more case studies.

%% file: 9-conclusion.tex
\section{Conclusion}

In this work, we introduce \bench{}: a comprehensive benchmark to evaluate the web page understanding and grounding capabilities of MLLMs.
\bench{} encompasses seven tasks spanning three different levels covering web page, element, and user action.
Unlike existing benchmarks, our benchmark aims to comprehensively evaluate MLLMs in web contexts, including understanding, OCR, grounding, and reasoning.
Our evaluation of 14 open-source MLLMs, Gemini Pro, Claude Sonnet, Claude Opus, and GPT-4V(ision) shows the substantial challenges posed by realistic web tasks.
Further analysis highlights several limitations of current MLLMs, including inadequate grounding in text-rich environments and subpar performance with low-resolution image inputs.
We believe \bench{} will serve as a catalyst for further exploration in the development of MLLMs towards artificial general intelligence.